\documentclass[wcp]{jmlr}

\usepackage{longtable}% for long tables

\usepackage{booktabs}
\usepackage{threeparttable}
\usepackage{multirow}

\jmlrvolume{95}
\jmlryear{2018}
\jmlrworkshop{ACML 2018}

\title[]{RDEC: Integrating Regularization into Deep Embedded Clustering for Imbalanced Datasets}
\newcommand{\argmax}{\mathop{\rm arg~max}\limits}

\usepackage{lscape}
\usepackage{slashbox}

  \author{\Name{Yaling Tao} \MakeLowercase{\Email{yaling1.tao@toshiba.co.jp}}\\
  \Name{Kentaro Takagi} \Email{kentaro1.takagi@toshiba.co.jp}\\
  \Name{Kouta Nakata} \Email{kouta.nakata@toshiba.co.jp}\\
  \addr Corporate R\&D Center, Toshiba Corporation, Kawasaki, Japan
 }

\editors{Jun Zhu and Ichiro Takeuchi}

\begin{document}

\maketitle

\begin{abstract}
Clustering is a fundamental machine learning task and can be used in many applications. With the development of deep neural networks (DNNs), combining techniques from DNNs with clustering has become a new research direction and achieved some success. However, few studies have focused on the imbalanced-data problem which commonly occurs in real-world applications. In this paper, we propose a clustering method, regularized deep embedding clustering (RDEC), that integrates virtual adversarial training (VAT), a network regularization technique, with a clustering method called deep embedding clustering (DEC). DEC optimizes cluster assignments by pushing data more densely around centroids in latent space, but it is sometimes sensitive to the initial location of centroids, especially in the case of imbalanced data, where the minor class has less chance to be assigned a good centroid. RDEC introduces regularization using VAT to ensure the model's robustness to local perturbations of data. VAT pushes data that are similar in the original space closer together in the latent space, bunching together data from minor classes and thereby facilitating cluster identification by RDEC. Combining the advantages of DEC and VAT, RDEC attains state-of-the-art performance on both balanced and imbalanced benchmark/real-world datasets. For example, accuracies are as high as 98.41\% on MNIST dataset and $85.45\%$ on a highly imbalanced dataset derived from the MNIST, which is nearly $8\%$ higher than the current best result.

\end{abstract}

\section{Introduction}
\label{Introduction}

Clustering is a fundamental machine learning method that groups data into clusters according to some measure of similarity or distance. It is commonly used to understand or summarize data about which we have no prior knowledge. Clustering can be widely utilized in practical applications, such as customer segmentation [\cite{ngai2009application}], text categorization [\cite{steinbach2000comparison}], genome analysis [\cite{sturn2002genesis}], intrusion prevention, and outlier detection [\cite{hodge2004survey}]. There are thus great needs to study the clustering problem. 

Improving the performance of clustering has been approached in many ways [\cite{estivill2002so}][\cite{berkhin2006survey}], such as by creating or refining algorithms that directly perform the  clustering task and by processing the data to make it more clustering-friendly. Traditional clustering algorithms, such as K-means, DBSCAN, Spectral Clustering, have been widely used for clustering analysis [\cite{estivill2002so}]. Dimensionality reduction and representation-learning techniques, such as principal component analysis (PCA) and non-negative matrix factorization (NMF) [\cite{lee2001algorithms}], have also been used extensively alongside clustering. In real-world applications, due to the diversity of datasets, careful selection of clustering algorithms and data processing techniques is required [\cite{liu2005toward}].

%trend of clustering 
With the development of deep learning techniques, the concept of {\it deep clustering}, which integrates deep neural networks (DNNs) with conventional clustering methods, has attracted considerable attention among researchers. For example, deep embedded clustering (DEC) [\cite{DEC}] defines an effective objective function as the KL divergence loss between the predicted distribution and an auxiliary target distribution of labels. Variational deep embedding (VaDE) [\cite{VaDE}] models a data generation procedure and picks clusters from Gaussian mixture models (GMM). Information maximizing self-augmented training (IMSAT) [\cite{IMSAT}] learns the distribution of labels by maximizing the information-theoretic dependency between data and their representations. Deep clustering using these and other methods has become increasingly widespread. 

However, few studies have focused on the imbalanced-data problem, which arises naturally in real-world applications. A dataset is said to be {\it imbalanced} when the numbers of data points belonging to different classes are significantly different, a common occurrence. Examples include the existence of rare diseases in medical diagnostics datasets, and the existence of rare defective products in production inspection datasets. Imbalance typically appears as a significant reduction in the performance attainable by most methods, which assume a relatively even label distribution [\cite{sun2007cost}]. This problem also exists in supervised and semi-supervised learning, where some studies have been performed. However, traditional approaches, such as re-sampling [\cite{chawla2004special}] and class-weighted cross-entropy [\cite{ronneberger2015u}], cannot be used in clustering because they require prior knowledge of the labels. There remains a need for an effective method that can improve performance for both balanced and imbalanced datasets.

From a theoretical analysis and preliminary experiments on the latest deep clustering methods, we find DEC to be a promising method that is relatively robust to imbalanced data. DEC optimizes clustering assignments by pushing data more densely around centroids in latent space. When each centroid is initialized at a location surrounded by similar data, DEC is expected to perform well regardless of whether the dataset is balanced or imbalanced.

However, DEC still has room for improvement with regard to class imbalance. In this work, we focus on two properties of DEC. First, DEC is sensitive to the initial location of centroids, which are randomly determined with K-means. With imbalanced data, in particular, minor classes are less likely to be assigned to good centroids, degrading DEC performance. We call this the {\it initial centroid problem}. Second, DEC tends to assign marginal data points (those far from all cluster centroids) to smaller clusters. Because it is unreasonable to set higher priorities to smaller clusters without knowing the class distributions, this rule may significantly degrade performance when applied to imbalanced data. We call this the {\it marginal data problem}. 

In this work, we apply virtual adversarial training (VAT) [\cite{VAT}] to mitigate these problems. VAT is a data augmentation technique originally proposed for supervised and semi-supervised learning. It aims to minimize difference in label distributions of input data and augmented data, the latter of which is generated by adding a small perturbation to the input data. The essential task of clustering is to gather similar data together, so VAT is in good agreement with clustering on this point since the augmented data can be interpreted as similar data. Our method thus uses VAT to augment the DEC loss function as a regularization term. We call our method regularized deep embedded clustering (RDEC). By integrating VAT, data points located near each other in the original space tend to be located together in the latent space, and data assembly considers not only centroids but also nearby data points. The contributions of this work are summarized as follows.
\begin{itemize} 
 \item We propose RDEC as a deep clustering method that improves conventional DEC in two ways: 1) by improving accuracy for the whole dataset despite centroids not being placed in particularly good locations during initialization, a common occurrence with imbalanced datasets, and 2) by improving accuracy for data near the margins of clusters.
 \item We conduct extensive experimental evaluation of deep clustering for comparison with current methods and analyse why RDEC works well. Our experimental results show that RDEC outperforms conventional methods on most benchmark datasets, particularly on imbalanced datasets. 
 \item We apply RDEC to a real-world application, namely clustering in wafer defect maps to find typical defective patterns in semiconductor manufacturing. Good performance on this dataset demonstrates RDEC's promising effectiveness for real-world applications. 
\end{itemize}

%%%%%%%%%%%%%%%%%%%%%%%%%%%%%%%%%%%%%%%%%%%%%%%%%%%%%%%%%%%%%%%%%%%%%%%%%%%%%%%%%%%%%%%%%%%%%%%%%%
\section{Related works} 
\label{RelatedWork} 
%AE+K-means
Most early deep clustering methods, such as [\cite{vincent2010stacked}] and [\cite{tian2014learning}], are two-stage methods that apply clustering after learning low-dimensional representations of the data in nonlinear latent space. The autoencoder method proposed in [\cite{hinton2006reducing}] is one of the most competitive methods used to learn feature representations.
However, since no cluster-promoting objective is explicitly incorporated into the learning process, the learned representations are not necessarily clustering-friendly.

%DEC, DCN, jointly learning .......
In view of this problem, representation learning and clustering are performed simultaneously in recent works such as [\cite{song2013auto}], [\cite{DEC}], and [\cite{DCN}]. However, these works do not sufficiently address the imbalanced-data problem. DEC [\cite{DEC}] exhibits some degrees of robustness to imbalanced datasets, but work is needed to further improve its robustness.
%############################################################# 
 
Several methods based on generative models have also been proposed [\cite{VaDE}], [\cite{dilokthanakul2016deep}].
VaDE [\cite{VaDE}] models a data generation procedure 
based on the variational autoencoder, where the data distribution in the latent space is modeled by GMM, the representations are sampled and then mapped into the space via the DNN.
This approach is novel and can work well in some cases. However, because the class distribution is unknown in imbalanced dataset, it is difficult to learn a good generative model, which may lead to low versatility and robustness.

%==============network regularization============== 
Regularization is an important research direction in deep learning. It is often used to avoid overfitting and to improve accuracy in supervised and semi-supervised learning [\cite{goodfellow2016deep}]. 
VAT was also used as a regularization term in IMSAT [\cite{IMSAT}], where the clustering term aims to maximize the information-theoretic dependency between data and predicted representations in an attempt to make representations in the latent space consistent with the original data. However, the maximum tends to be obtained when all clusters are uniform, meaning it does not perform well on imbalanced datasets.
  
\section{Regularized deep embedded clustering}
\label{Model}
\subsection{Notation}
Consider a dataset $X$ consisting of $n$ data vectors with dimensionality $d$. Let $x_i\in R^d$ denote item $i$ in $X$ (index $i$ can be omitted for simplicity when there is no need for specificity), and let $K$ denote the number of clusters, which is assumed to come from prior knowledge. Clusters are indexed from $0$ to $K-1$, and each cluster is represented by a centroid $u_j (j=0,1,...,K-1)$. Our task is to assign items $x$ into $K$ clusters. Instead of clustering directly in the data space $X$, data points are represented in a latent space $Z$ via a nonlinear mapping $f_\theta: X \rightarrow Z$, where $\theta$ is a set of learnable parameters. DNN is used to parametrize $f_\theta$. 
 
\begin{figure}[htp]
\begin{center} 
\includegraphics[width=0.4\textwidth]{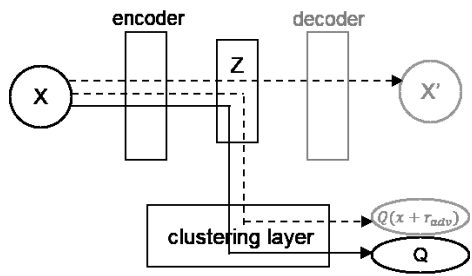}
\caption{RDEC Model.}\label{fig:RdecModel}
\end{center}
\end{figure}

Figure \ref{fig:RdecModel} shows the RDEC network model, which comprises three sub-models. That connected by solid lines is the main clustering model, where data $x$ are mapped into embedded representations $z$, and then transformed into their predicted distributions via a clustering layer. The clustering layer is the same as that of DEC.
The other two sub-models connected by dotted lines are the autoencoder and VAT models. The autoencoder model is used for initialization and the VAT model for regularization. Both promote the clustering model. The learning process is divided into two phases: pretraining with autoencoder model and fine-tuning with the clustering and VAT models. 
Because the decoder is not used after pretraining and the VAT output is not needed in our final results, they are grayed out in Figure \ref{fig:RdecModel}.
We describe the clustering model and VAT model separately below. Their details can be found in [\cite{DEC}] and [\cite{VAT}], respectively.

\subsection{Clustering model} 
Clustering is performed in the latent space $Z$. First, embedded data $z_i$ is assigned to cluster $u_j$ with probability $q_{ij} (q_{ij}\in Q)$.
%Each $q_{ij}$ is the similarity between $z_i$ and $u_j$ measured by Student's {\it t-}distribution as
Each $q_{ij}$ is the similarity between $z_i$ and $u_j$ measured by Student's {\it t-}distribution [\cite{tsne}] as
\begin{eqnarray} 
q_{ij}= \frac{(1+||z_i-u_j||^2/\alpha)^{-\frac{\alpha+1}{2}}}{\sum_{j'}(1+||z_i-u_{j'}||^2/\alpha)^{-\frac{\alpha+1}{2}}},
\label{eq:Q}
\end{eqnarray}
where $\alpha$ is the degree of freedom, set to 1 in this model. Centroids $u_j (j=0,1,...,K-1)$ are initialized with K-means on the embedded data $z$. 
Elements $q_{ij}$ of $Q$ are also called soft assignments, and Q is called predicted distribution of labels. 
An auxiliary target distribution $P$ corresponding to Q is defined, with each $p_{ij}\in P$ computed as
\begin{eqnarray} 
p_{ij}= \frac{q_{ij}^s/f_j}{\sum_{j'}q_{ij'}^s/f_{j'}},
\label{eq:P}
\end{eqnarray}
where $f_j =\sum_iq_{ij}$ are the soft cluster frequencies and $s$ is a constant. The original DEC fixed $s=2$, but we found that it works well in a larger range, e.g. $s\geq1$. In this definition, $q_{ij}$ is raised to the power of $s$. Compared with the predicted distribution $Q$, the bell-shaped curve of target distribution $P$ has a higher peak and lower tails. The numerator $q_{ij}^s$ aims to attract data toward centroids. When the initial centroids are placed unfavourably, the initial centroid problem described in Section \ref{Introduction} occurs. The denominator $f_j$ has a large effect on marginal data points that are far from all centroids. When $q_{ij}$ is nearly equal over $j$, $p_{ij}$ corresponding to the lower frequency $f_j$ will become higher and marginal data will be assigned to smaller clusters.
This dependency on $f_j$ is ill-suited to imbalanced datasets and will lead to the marginal data problem described in Section \ref{Introduction}.
 
The clustering model is trained by matching soft assignments to the target distribution.
The objective function is defined as a KL divergence loss between $Q$ and $P$ as
\begin{eqnarray} 
L_D=KL[P\parallel Q] = \sum_{i}\sum_{j}p_{ij}log\frac{p_{ij}}{q_{ij}}.
\label{eq:LD}
\end{eqnarray}

\subsection{VAT model}
VAT uses data augmentation to impose the intended invariance on the label distribution.
It enables similar data in the original space to follow similar distributions in the latent space.
The objective function is defined as a KL divergence loss between predicted distributions of $x$ and augmented data $x+r_{adv}$ as
\begin{eqnarray} 
L_V=KL[Q\parallel Q(x+r_{adv})],
\label{eq:LV}
\end{eqnarray}
where $r_{adv}$ is adversarial perturbation computed in an adversarial way as
\begin{eqnarray} 
r_{adv}= \argmax_{r;||r||\leq\epsilon}KL[Q||Q(x+r)],
\label{eq:radv}
\end{eqnarray}
where $r$ is a perturbation that does not alter the meaning of the data point and $\epsilon$ is the perturbation size, which is a hyper-parameter. The impact of $\epsilon$ is discussed in Section \ref{impactEpsilon}.
 
\subsection{Objective of RDEC}
From the above, the objective function of RDEC can be written as
\begin{eqnarray} 
L&= &L_{D}+\gamma L_{V}\\
  & = & KL[P\parallel Q] + \gamma KL[Q\parallel Q(x+r_{adv})],
\label{eq:loss}
\end{eqnarray}
where $\gamma\geq0$ is a weight that controls the balance of $L_{D}$ and $L_{V}$ in the loss function. Regularization with VAT reduces DEC's dependency on centroids and soft cluster frequencies by considering local similar data. The effects of VAT and $\gamma$ are discussed in Sections \ref{effectVAT} and \ref{effectGamma}, respectively.

The objective function in Eq.~(\ref{eq:loss}) is optimized using mini-batch stochastic gradient decent (SGD) and backpropagation. Latent representations $z_i$, cluster centroids $u_j$, and soft assignments $q_{ij}$ are updated at each iteration, while the target distribution $P$ is updated at an interval of  $\tau$ iterations. 
The learning procedure is stopped when the rate of changes in assignments between two consecutive iterations falls below a threshold $\sigma$ or the maximum number of iterations $Itr_{max}$ is reached. Because $L_V$ is learned in an adversarial way, RDEC would take triple the computation to update network parameters as compared with DEC. However, updating target distribution $P$ accounts for the main computation in both RDEC and DEC, so the gap in computation times is not large. The time complexities of DEC and RDEC in each iteration can be respectively written as  $O(\frac{1}{\tau}n)+O(b)$ and  $O(\frac{1}{\tau}n)+O(b')$, where $n$ is the number of samples, $b$ is the mini-batch size, and $b'=3b$.

\section{Experiments on benchmark datasets}
\label{Experiments}
We quantitatively compared the performance of RDEC with the performance of several baseline methods on a set of benchmark datasets.
The baseline methods were K-means, AE+K-means, AE+DBSCAN, DEC, IMSAT, VaDE, and DCN. AE+K-means and AE+DBSCAN are two-stage methods that execute K-means and DBSCAN after pretraining on latent features.  
Adjusted rand index (ARI) [\cite{yeung2001details}] and unsupervised clustering accuracy(ACC) are used as performance metrics. Following [\cite{DEC}], $ACC=\max_{\substack{m}}\frac{\sum_{i=1}^{n}1\{l_i=m(c_i)\}}{n}$, where $l_i$ is the ground-truth label, $c_i$ is the cluster assignment, and $m$ ranges over all possible one-to-one mappings between clusters and labels.

\subsection{Datasets}
Three benchmark datasets, MNIST, STL,  and Reuters, are used in our experiments. Each has a relatively balanced label distribution. Because we focus on the performance of imbalanced datasets, we generated imbalanced versions of these datasets for comparison. To describe the degree of imbalance, we used a metric named minimum retention rate $r_{min}$, defined as the ratio of disproportion between the numbers of samples from the minority and majority classes in the dataset. 

\begin{itemize} 
 \item MNIST: A dataset consisting of $70,000$ handwritten digits $0$ to $9$. All classes contain nearly equal numbers of samples [\cite{lecun1998mnist}]. Each digit is size-normalized as a $28\times28$-px image, represented as a 784-dimensional vector.
 \item MNIST-Imb-0: A variant of MNIST generated by reducing the number of 0-labeled samples to $1/10$ its original value. The $r_{min}$ of MNIST-Imb-0 is thus $0.1$.
 \item MNIST-Imb-all: Another variant of MNIST where the numbers of samples in the ten classes differ significantly. The respective numbers of samples from class $0$ to $9$ were $10$, $30$, $50$, $1,000$, $200$, $500$, $300$, $6,000$, $4,000$, and $800$, so the $r_{min}$ is $1/600$.
 \item STL: A set of $96\times96$-px color images [\cite{STL}]. There are ten classes labeled as airplane, bird, car, and so on. Each class contains $1,300$ images and each image is represented as a $27,648$-dimensional vector. 
 \item STL-VGG: $2,048$-dimensional feature vectors of STL, extracted using the VGG-16 model with weights pretrained on ImageNet [\cite{VGG16}].
 \item STL-VGG-Imb: A STL-VGG variant where the number of samples in one class is reduced to $1/10$, so the $r_{min}$ of this dataset is $0.1$.
 \item Reuters: A document dataset consisting of news stories labeled with a category tree [\cite{lewis2004reuters}]. Following [\cite{DEC}], four root categories are used as labels and uniquely labeled documents are extracted. Each article is represented as a feature vector using the tf-idf method on the $2,000$ most frequent words. The number of articles from the four classes are $40,635$, $25,457$, $22,356$, and $8,502$, respectively.
 \item Reuters-Imb: A variant of Reuters, generated by reducing the number of samples of the minority class to meet $r_{min}=0.1$.
 \item Reuters-10K: A random subset of $10,000$ documents sampled from Reuters. The number of articles from the four classes are $4,022$, $2,703$, $2,380$, and $895$, respectively.
 \end{itemize}

\subsection{Parameter settings}
Following [\cite{DEC}], we used fully connected networks with dimensions $d-500-500-2000-10$ for the encoder and $10-2000-500-500-d$ for the decoder, where $d$ is the input data dimension. Except for the input, output, and embedding layers, all internal layers are activated using the ReLU nonlinearity function. All datasets used the same network settings.

Rather than greedy layer-wise training as in DEC, in the pretraining phase, we directly trained the autoencoder. Our preliminary experiments showed that direct training works well at a lower cost. The optimizer SGD with learning rate $lr=1$ and momentum $\beta=0.9$ was used for MNIST and Reuters, and
the optimizer Adam [\cite{kingma2014adam}] with learning rate $\lambda = 0.001$, $\beta_1 = 0.9$, $\beta_2 = 0.999$ was used for STL. 
Pretraining epochs for MNIST, Reuters, and STL were set to 300, 50, and 10, respectively. 

In the fine-tuning phase, the clustering network was initialized with the trained encoder. 
For all datasets, the optimizer SGD with learning rate $lr=0.01$ and momentum $\beta=0.9$ was used, the maximum number of iterations $Itr_{max}$ was set to $20,000$, and the convergence threshold $\sigma$ was set to $0.01$. The update intervals $\tau$ for MNIST, Reuters, and STL were respectively set to $140$, $30$, and $30$.
Unless specifically stated otherwise, in these experiments parameters $\gamma=5$ and $s=2$. For VAT-related parameters, perturbation size $\epsilon=1$, mesh size $\xi=10$, and power iterations $ip=1$. 
In both the pretraining and finetuning phases, the mini-batch size was 256. All parameter settings of variant datasets were the same as in their original datasets.
All experiments were repeated five times and the means of ARI and ACC were used for comparison, and the standard deviations were also recorded for reference. 
Note that because two parameters of DBSCAN must be tuned depending on datasets, we used a grid search method to choose appropriate values for each dataset.

\subsection{Results}
\label{results}

Table \ref{tab:ret} shows the clustering results of the baseline methods compared with RDEC. From these results, we can see that RDEC outperforms other methods for most datasets, according to both metrics ACC and ARI. In particular, RDEC yielded a high accuracy of $98.41\%$ on MNIST with a small standard deviation of $0.01$, which is comparable with supervised and semi-supervised methods. 
IMSAT also performed well on MNIST, with small gaps in ACC and ARI between it and RDEC. However, for the imbalanced MNIST, gaps exceeded $20\%$ in the case of MNIST-Imb-all. The results also validated our analysis of IMSAT in Section \ref{RelatedWork}. Note that because the metric ACC is in good agreement with ARI in our results, following previous works such as DEC, in the following we reserve ACC for only comparison. 

\begin{table}[htb]
\begin{center}
\caption{Clustering results on various benchmark datasets (\%).} 
\scalebox{0.7}{
\begin{tabular}[c]{c|c|c|c|c|c|c|c|c|c}\hline
 \backslashbox{\scriptsize{Dataset}}{\scriptsize{Method}} &&K-means&\shortstack{AE+\\K-means}&\shortstack{AE+\\DBSCAN}&DEC&IMSAT&VaDE&DCN&\shortstack{RDEC}\\\hline\hline

 \multirow{2}{*}{MNIST}&ACC&53.38\scriptsize{$\pm0.13$}&89.93\scriptsize{$\pm2.43$} & 47.67\scriptsize{$\pm0.14$} &93.98\scriptsize{$\pm2.80$}&97.76\scriptsize{$\pm0.65$}&74.05 \scriptsize{$\pm3.01(\ast1)$}&54.15\scriptsize{$\pm1.25(\ast5)$}& {\bf 98.41}\scriptsize{$\pm0.01$}\\\cline{2-10}
 &ARI&36.61\scriptsize{$\pm0.08$}&80.25\scriptsize{$\pm3.03$} &17.10\scriptsize{$\pm0.16$} & 88.11\scriptsize{$\pm4.59$}&95.15\scriptsize{$\pm1.34$}&65.80 \scriptsize{$\pm4.94$}&38.64\scriptsize{$\pm1.25$}&{\bf 96.51}\scriptsize{$\pm0.02$}\\ \hline

\multirow{2}{*}{MNIST-Imb-0}&ACC& 50.44\scriptsize{$\pm0.07$}& 86.52\scriptsize{$\pm1.23$} &47.38\scriptsize{$\pm0.14$} & 89.85\scriptsize{$\pm1.03$} & 86.72\scriptsize{$\pm7.24$} & N/A & 50.74\scriptsize{$\pm0.37$}  &{\bf 94.82} \scriptsize{$\pm1.48$}\\\cline{2-10}
&ARI& 33.49\scriptsize{$\pm0.02$}& 78.78\scriptsize{$\pm1.52$} &18.32\scriptsize{$\pm0.54$} & 86.30\scriptsize{$\pm1.60$} & 83.65\scriptsize{$\pm9.59$} & N/A & 34.06\scriptsize{$\pm0.83$}  & {\bf93.51} \scriptsize{$\pm0.75$}\\\hline

\multirow{2}{*}{MNIST-Imb-all}&ACC& 29.19\scriptsize{$\pm0.20$}& 45.68\scriptsize{$\pm1.18$} & 38.64\scriptsize{$\pm0.00$} & 40.71\scriptsize{$\pm1.43$} & 32.09\scriptsize{$\pm1.19$} & N/A & 40.59\scriptsize{$\pm0.53$}  &  {\bf  65.40}\scriptsize{$\pm5.33$}  \\\cline{2-10}
&ARI& 19.56\scriptsize{$\pm0.02$}& 33.36\scriptsize{$\pm1.58$} & 0.01\scriptsize{$\pm0.00$} & 30.62\scriptsize{$\pm2.25$} & 24.39\scriptsize{$\pm1.39$} & N/A & 30.21\scriptsize{$\pm1.92$} &  {\bf  53.93}\scriptsize{$\pm3.80$}  \\\hline\hline

\multirow{2}{*}{Reuters-10K}&ACC& 57.48\scriptsize{$\pm7.69$} & 60.68\scriptsize{$\pm5.29$} & 37.47\scriptsize{$\pm0.23$} & 66.48\scriptsize{$\pm4.20$} & {\bf 72.34}\scriptsize{$\pm3.70$} & 56.73\scriptsize{$\pm2.99(\ast2)$} &N/A & 66.18\scriptsize{$\pm4.70$}\\\cline{2-10}
&ARI& 31.25\scriptsize{$\pm7.61$} & 28.79\scriptsize{$\pm5.59$} &  1.31\scriptsize{$\pm1.60$} & 36.51\scriptsize{$\pm7.54$} & {\bf 45.77}\scriptsize{$\pm6.52$} & 25.90\scriptsize{$\pm3.06$} &N/A & 37.01\scriptsize{$\pm8.41$}\\\hline

\multirow{2}{*}{Reuters}&ACC& 53.33\scriptsize{$\pm0.27$} & 74.17\scriptsize{$\pm2.53$} & 34.22\scriptsize{$\pm0.14$} & 73.88\scriptsize{$\pm3.65$} & 67.78\scriptsize{$\pm5.25$} & 43.69 \scriptsize{$\pm0.01(\ast3)$} &N/A &  {\bf75.45} \scriptsize{$\pm3.06$} \\\cline{2-10}
&ARI& 25.43\scriptsize{$\pm2.96$} & 56.78\scriptsize{$\pm6.74$} & 1.79\scriptsize{$\pm0.37$} & 57.58\scriptsize{$\pm3.47$} & 41.97\scriptsize{$\pm5.18$} &N/A&N/A &  {\bf61.48} \scriptsize{$\pm2.47$} \\\hline 

\multirow{2}{*}{Reuters-Imb}&ACC& 55.58\scriptsize{$\pm0.03$} & 72.89\scriptsize{$\pm4.18$} & 32.52\scriptsize{$\pm0.01$}  &71.85\scriptsize{$\pm3.45$} & 68.12\scriptsize{$\pm2.91$} &N/A&N/A &{\bf 73.87}\scriptsize{$\pm3.50$}\\\cline{2-10}
&ARI&29.99\scriptsize{$\pm0.05$} & 50.01\scriptsize{$\pm8.72$} &0.01\scriptsize{$\pm0.00$}  &51.40\scriptsize{$\pm8.28$} & 42.61\scriptsize{$\pm3.57$} &N/A&N/A &{\bf 55.55}\scriptsize{$\pm9.25$}\\\hline\hline

\multirow{2}{*}{STL-VGG}&ACC&72.66\scriptsize{$\pm5.37$} &77.93\scriptsize{$\pm5.25$} &15.01\scriptsize{$\pm0.89$} &84.37\scriptsize{$\pm5.06$}&83.30\scriptsize{$\pm6.04$} &\scriptsize{$(\ast4)$}  & 80.44\scriptsize{$\pm6.56$} &{\bf86.59}\scriptsize{$\pm7.48$}\\\cline{2-10}
&ARI&55.24\scriptsize{$\pm4.45$} &62.79\scriptsize{$\pm3.84$} &2.03\scriptsize{$\pm1.00$} &75.89\scriptsize{$\pm3.61$}&72.80\scriptsize{$\pm4.78$} &N/A& 64.54\scriptsize{$\pm5.46$} & {\bf78.24} \scriptsize{$\pm7.20$}\\\hline

\multirow{2}{*}{STL-VGG-Imb}&ACC& 76.70\scriptsize{$\pm3.13$}  & 82.08\scriptsize{$\pm1.86$} &16.93\scriptsize{$\pm1.00$} & 87.21\scriptsize{$\pm2.46$}  & 81.30\scriptsize{$\pm2.29$}  &N/A& 80.74\scriptsize{$\pm2.12$}  &{\bf 88.63}\scriptsize{$\pm0.14$}  \\\cline{2-10}
&ARI& 60.37\scriptsize{$\pm1.75$}  & 69.06\scriptsize{$\pm3.06$} &2.42\scriptsize{$\pm1.11$} &78.84\scriptsize{$\pm3.55$}  &72.87\scriptsize{$\pm2.01$}  &N/A& 66.56\scriptsize{$\pm1.59$} &{\bf 80.56}\scriptsize{$\pm0.26$}  \\\hline

\multirow{2}{*}{STL}&ACC& 21.98\scriptsize{$\pm0.04$}  & 22.64\scriptsize{$\pm1.49$} & 10.00\scriptsize{$\pm0.00$}  & 21.04\scriptsize{$\pm0.77$}  & {\bf 24.70}\scriptsize{$\pm0.73$}  &N/A&N/A  &  21.27\scriptsize{$\pm0.90$}  \\\cline{2-10}
&ARI& 6.03\scriptsize{$\pm0.01$}  & 6.39\scriptsize{$\pm0.66$} & 0.00\scriptsize{$\pm0.00$}  & 5.98\scriptsize{$\pm0.50$}  & {\bf 8.45}\scriptsize{$\pm0.26$}  &N/A&N/A &  5.90\scriptsize{$\pm0.53$}  \\\hline\hline 

\multicolumn{10}{c}{$\scriptsize{(\ast1)}: 94.46$, $\scriptsize{(\ast2)}: 79.83$, $\scriptsize{(\ast3)}: 79.38$, $\scriptsize{(\ast4)}: 84.45$ , $\scriptsize{(\ast5)}: 83.00$   }\\ 
\multicolumn{10}{l}{Note: Because we could not reproduce the accuracies of VaDE and DCN reported in the original papers using their authors'}\\
\multicolumn{10}{l}{code (obtained from github), we report our results in this table and note the authors' results in footnotes.}
\end{tabular}
}
\label{tab:ret}
\end{center}
\end{table}

RDEC clearly performed better than did DEC. For example, RDEC improved ACC on MNIST and MNIST-Imb-all by nearly $5\%$ and $25\%$, respectively. These results reflect the effect of VAT. Table \ref{tab:ret} also shows that all methods yielded poor performance for STL. For this difficult dataset, other techniques need to be combined with clustering. The results of STL-VGG demonstrated the feasibility of this idea. For example, VGG-16 significantly increased the accuracy on STL-VGG. Since the weights in VGG-16 are trained on ImageNet, from which STL was acquired, it is reasonable that STL-VGG yielded good performance. These results suggest that pretrained weights obtained from a dataset, those similar to the provided dataset but with easier-to-obtain labels, can be used for the current dataset.

\subsection{Performance on imbalanced datasets}
\label{perfomanceImb}
To further examine the effect of our method on imbalanced datasets, following the experimental method of DEC [\cite{DEC}], we sampled subsets of MNIST with various retention rates. For minimum
retention rate $r_{min}$, class-0 data points are kept at a probability of $r_{min}$ and those for class 9 with a probability of 1, with the other classes retained according to linear interpolation between. 
The results shown in Table \ref{tab:retImb} also demonstrate the superiority of RDEC. Even for the most imbalanced subset with $r_{min} =0.1$, ACC obtained by RDEC is higher than $85\%$.

\begin{table}[htb]
\begin{center}
\caption{Clustering accuracy on imbalanced subsample of MNIST (\%).} 
  \begin{tabular}{c|c|c|c|c|c}\hline
  \backslashbox{\scriptsize{Method}}{\scriptsize{$r_{min}$}}&0.1&0.3&0.5&0.7&0.9\\\hline\hline
   K-means& 46.30\scriptsize{$\pm0.07$} & 51.40 \scriptsize{$\pm2.40$} & 53.52\scriptsize{$\pm0.04$}  & 53.88\scriptsize{$\pm0.02$}  & 54.33\scriptsize{$\pm0.02$} \\\hline
  AE+K-means& 76.97\scriptsize{$\pm2.74$} & 86.19\scriptsize{$\pm3.40$} & 88.44\scriptsize{$\pm02.42$} & 90.62\scriptsize{$\pm0.32$} & 86.58\scriptsize{$\pm4.29$}\\\hline
  DEC& 77.04\scriptsize{$\pm2.85$}& 89.53\scriptsize{$\pm4.32$}& 92.75\scriptsize{$\pm3.22$}& 95.24\scriptsize{$\pm0.38$}& 89.55\scriptsize{$\pm5.13$}\\\hline
   RDEC& {\bf 85.45}\scriptsize{$\pm0.86$}& {\bf 96.03}\scriptsize{$\pm4.24$}& {\bf 98.11}\scriptsize{$\pm0.01$}& {\bf 98.30}\scriptsize{$\pm0.03$}& {\bf 96.02}\scriptsize{$\pm5.27$}\\\hline
 \end{tabular}
 \label{tab:retImb}
 \end{center}
\end{table}
 
To investigate the performance on the minority class, we also calculated the recall, precision, and F-measure values of class 0, shown in Table \ref{tab:Fmeasure}. It is evident that the results as evaluated by recall, precision, and F-measure are in good agreement with those above as evaluated by ACC. RDEC indeed improves clustering performance on imbalanced datasets. 
\begin{table}[h]
\begin{center}
\caption{F-measures for the 0-class of imbalanced versions of MNIST (\%).}
\scalebox{0.9}{
  \begin{tabular}{c|c|c|c|c|c|c}\hline
 \backslashbox{\scriptsize{Method}}{\scriptsize{$r_{min}$}}  & &0.1&0.3&0.5&0.7&0.9\\\hline\hline
\multirow{3}{*}{K-means}&Recall& 55.74\scriptsize{$\pm0.08$} & 81.30\scriptsize{$\pm1.52$} & 77.76\scriptsize{$\pm0.03$}& 81.13\scriptsize{$\pm0.01$}& 80.02\scriptsize{$\pm0.05$}\\ \cline{2-7}
            &Precision& 12.47\scriptsize{$\pm0.03$} & 84.23\scriptsize{$\pm0.23$} & 90.69\scriptsize{$\pm0.04$}& 91.98\scriptsize{$\pm0.03$}& 93.38\scriptsize{$\pm0.01$}\\ \cline{2-7}
            &F-measure& 20.39\scriptsize{$\pm0.05$} & 82.73\scriptsize{$\pm0.68$} & 83.73\scriptsize{$\pm0.00$}& 86.21\scriptsize{$\pm0.00$}& 86.18\scriptsize{$\pm0.03$}\\\hline

\multirow{3}{*}{AE+K-means}&Recall& 67.13\scriptsize{$\pm26.82$} & 97.10\scriptsize{$\pm0.90$} & 96.66\scriptsize{$\pm1.00$}& 97.68\scriptsize{$\pm0.28$}& 97.79\scriptsize{$\pm1.12$}\\ \cline{2-7}
          &Precision& 29.67\scriptsize{$\pm19.69$} & 83.57\scriptsize{$\pm2.43$} & 88.63\scriptsize{$\pm2.07$}& 92.37\scriptsize{$\pm0.79$}& 93.86\scriptsize{$\pm0.99$}\\ \cline{2-7}
          &F-measure& 40.58\scriptsize{$\pm23.83$} & 89.81\scriptsize{$\pm1.43$} & 92.46\scriptsize{$\pm1.33$}& 94.95\scriptsize{$\pm0.46$}& 95.79\scriptsize{$\pm1.04$}\\\hline

\multirow{3}{*}{DEC}&Recall& 78.90\scriptsize{$\pm22.69$} & 97.86\scriptsize{$\pm0.23$} & 98.10\scriptsize{$\pm0.16$}& 98.40\scriptsize{$\pm0.11$}& 98.58\scriptsize{$\pm0.10$}\\ \cline{2-7}
            &Precision& 26.77\scriptsize{$\pm12.97$} & 88.99\scriptsize{$\pm2.52$} & 94.18\scriptsize{$\pm0.92$}& 96.71\scriptsize{$\pm0.55$}& 97.30\scriptsize{$\pm0.69$}\\ \cline{2-7}
            &F-measure& 39.64\scriptsize{$\pm17.14$} & 93.20\scriptsize{$\pm1.35$} & 96.10\scriptsize{$\pm0.49$}& 97.55\scriptsize{$\pm0.26$}& 97.93\scriptsize{$\pm0.36$}\\\hline

\multirow{3}{*}{RDEC}&Recall& {\bf98.20}\scriptsize{$\pm0.39$} & {\bf98.92}\scriptsize{$\pm0.14$} & {\bf99.33}\scriptsize{$\pm0.05$}& {\bf 99.42}\scriptsize{$\pm0.11$}& {\bf 99.58}\scriptsize{$\pm0.10$}\\ \cline{2-7}
            &Precision& {\bf 60.57}\scriptsize{$\pm28.66$} & {\bf 96.53}\scriptsize{$\pm0.26$} & {\bf 98.13}\scriptsize{$\pm0.18$}& {\bf 98.38}\scriptsize{$\pm0.13$}& {\bf 98.74}\scriptsize{$\pm0.13$}\\ \cline{2-7}
            &F-measure& {\bf 71.77}\scriptsize{$\pm22.21$} & {\bf 97.71}\scriptsize{$\pm0.16$} & {\bf98.72}\scriptsize{$\pm0.07$}&  {\bf98.89}\scriptsize{$\pm0.08$}& {\bf 99.16}\scriptsize{$\pm0.05$}\\\hline
 \end{tabular}
}
 \label{tab:Fmeasure}
 \end{center}
\end{table}

There are two reasons why RDEC performs well on imbalanced datasets. First, DEC inherently emphasizes cluster purity, improving robustness against imbalanced data to some degree. In addition, VAT makes up for the weaknesses of DEC, allowing RDEC to significantly outperform DEC.  

In DEC, all assignments are performed on the basis of distances between embedded points and centroids, which are independent of cluster size. From the definition of the target distribution $P$ in Eq.~(\ref{eq:P}), we can see that soft assignments $q$ are raised to the power of $s$ and then divided by cluster frequency $f$, so data close to a centroid will move more closely toward that centroid, and data far from all cluster centroids will move to the one with less samples in the learning process. The whole learning process makes sure that data points assemble around the centroids. DEC performs well if the location of centroids is proper to the dataset and soft assignments have a relatively high level of confidence.

RDEC alleviates the initial centroid and marginal data problems. 
By jointly optimizing $L_D$ and $L_V$, data points are not only drawn by their centroids but also grabbed by other nearby points. 
This reduces centroid dependency and makes marginal data select centroid following data near them.
This hypothesis is confirmed in the following Section \ref{effectVAT}. 

\subsection{Effect of VAT}
\label{effectVAT}
We investigated the effects of VAT on balanced and imbalanced datasets using two subsets of MNIST, one consisting of all 0-labeled and 6-labeled samples, and another consisting of 1/10 of 0-labeled and all 6-labeled samples.
To visualize the learning process, the dimension of the embedded layer was set to $2$.
DEC and RDEC were compared in both experiments, with
the seed value fixed to guarantee that the two methods learned under the same conditions.

\begin{figure}[htb]
 \begin{center}
  \subfigure[]{	
   \includegraphics[width=1\textwidth]{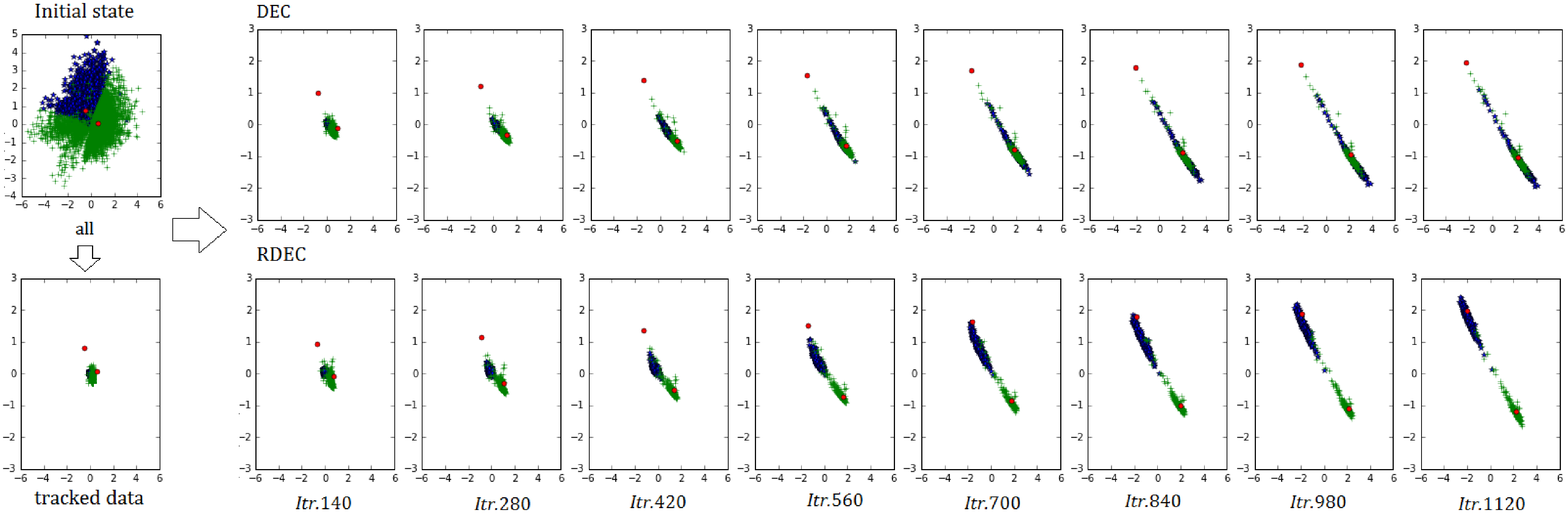}\label{fig:process}
  }\\
  \subfigure[]{
   \includegraphics[width=1\textwidth]{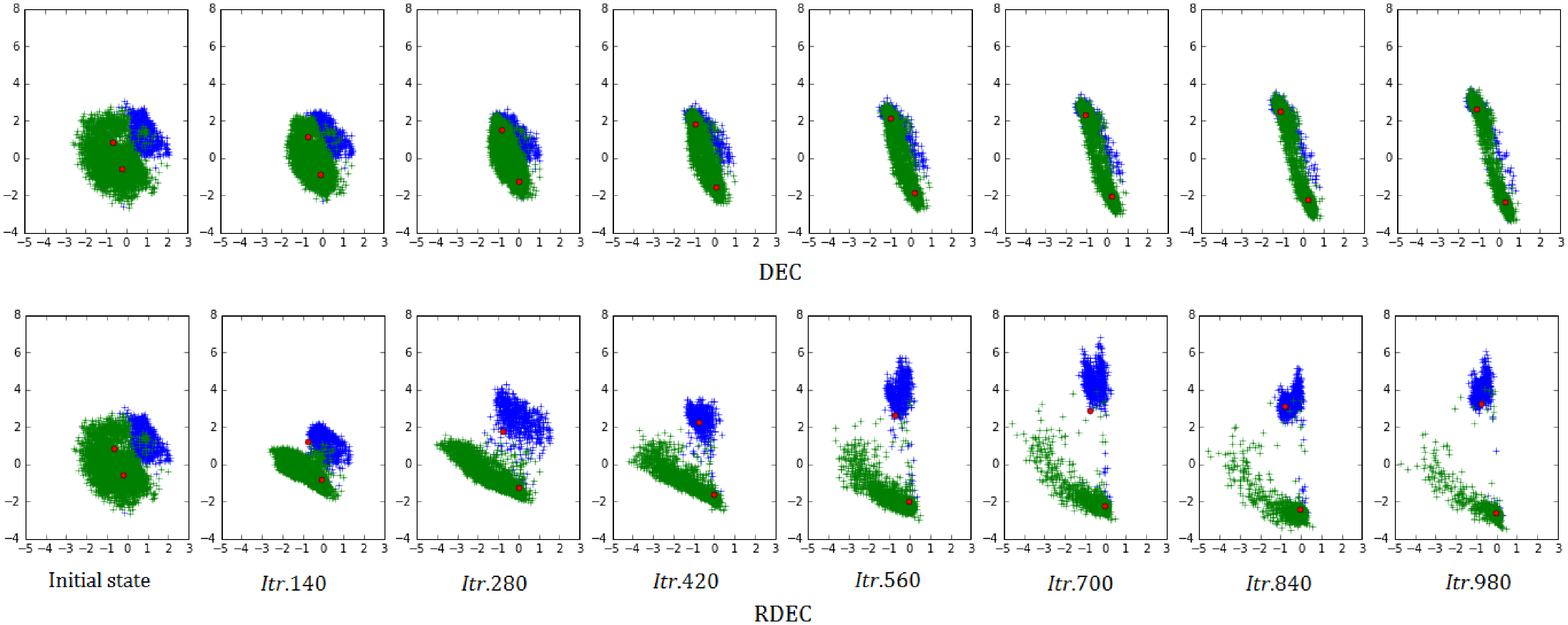}\label{fig:processImb}
  }  
 \caption{Visualized learning processes for two MNIST subsets: a) a balanced dataset consisting of all 0-labeled and 6-labeled data; 
b) an imbalanced dataset consisting of 1/10 of 0-labeled and all 6-labeled data.} 
  \label{Fig:pro}
 \end{center}
\end{figure}

Figure \ref{fig:process} illustrates the learning processes on the balanced MNIST subset. The first column displays the initial state where data points are embedded by autoencoder and centroids are initialized by K-means. Centroids and 0-labeled and 6-labeled data points are respectively colored red, blue, and green. To improve visibility, only a small number of points (about $280$ 0-labeled and $220$ 6-labeled points around one centroid) were selected to be tracked in the following learning processes. The leftmost panel in the first row displays the whole data, and the panel subfigure in the second row displays the tracked data. From the second column, the first row displays the DEC learning process and the second row displays that of RDEC, thereby visualizing processes at intervals of 140 iterations. 
 
Both the DEC and RDEC learning processes show clear trends in which data points assemble around the centroids. This reflects the fact that both methods optimize the $L_D$ in Eq.~(\ref{eq:LD}) in the same way. More importantly, data movements in RDEC are significantly clustering-friendly, with similar data joined and attracted toward the centroid as a whole. This phenomenon validates the hypothesis in Section \ref{perfomanceImb}. This property of VAT is very useful, especially for data near margins, where centroid selection has a low level of confidence. The accuracies  of DEC and RDEC in this case are $84.82\%$ and $98.29\%$, respectively.

Figure \ref{fig:processImb} illustrates learning processes on the imbalanced MNIST subset. 
The effect of VAT is more significant in the imbalanced scenario. 
The performance of DEC was not very good in this case. Although the data points assemble around centroids just as well as in the example shown in Figure \ref{fig:process}, the less-favourable centroid location divided data points from the same class by a perpendicular bisector line between two centroids, resulting in impure clusters.
The RDEC learning process, in contrast, demonstrated strong robustness to this centroid location. Data movements by VAT are more significant than those by centroids.
The accuracies of DEC and RDEC in this case are $61.57\%$ and $99.37\%$, respectively.

\subsection{Effect of parameter $\gamma$}
\label{effectGamma}
\begin{figure}[htbp]
 \begin{tabular}{cc}
 \begin{minipage}{0.333\textwidth}
 \begin{flushleft}
 \includegraphics[width=0.92\textwidth]{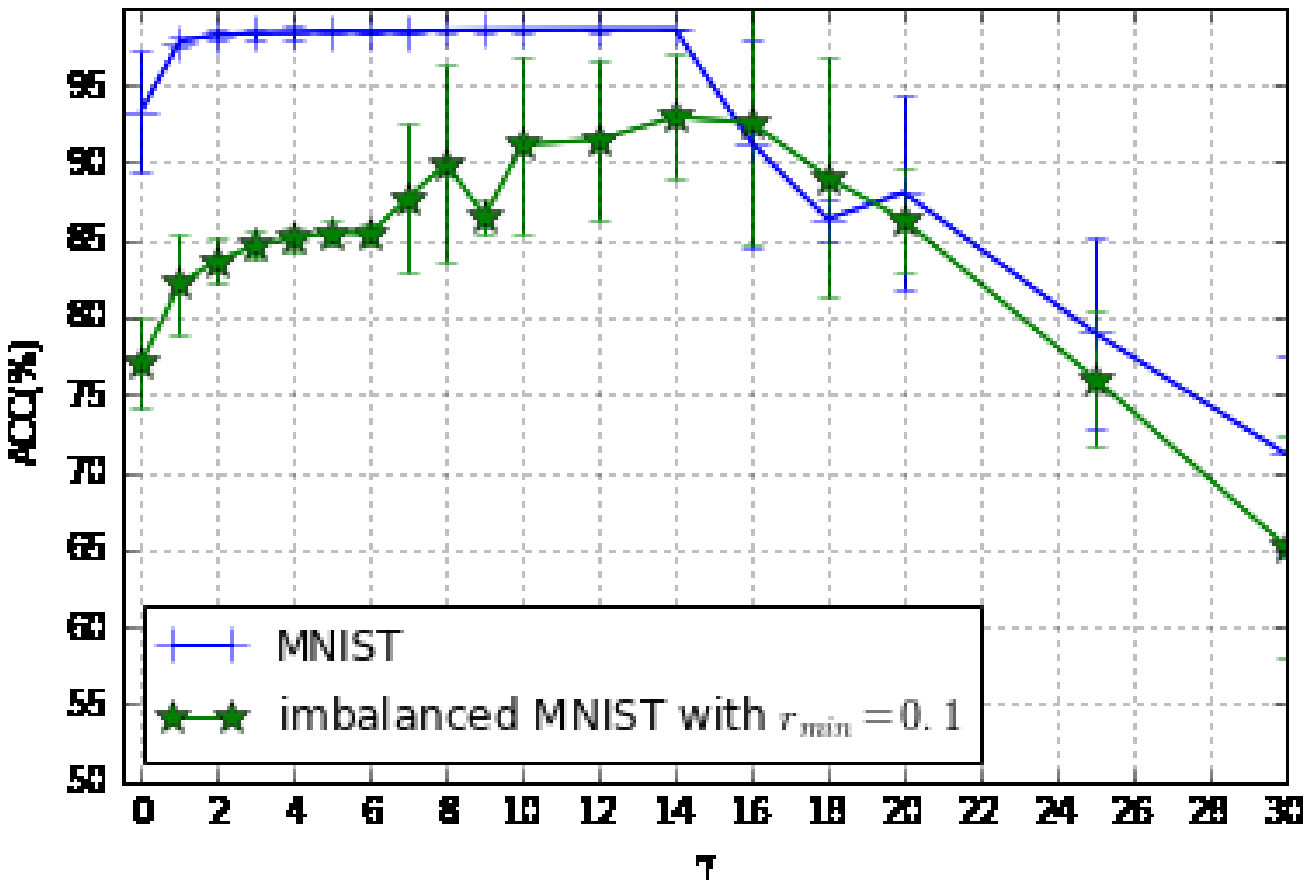}
 \caption{\footnotesize Effect of parameter $\gamma$.}
 \label{fig:gamma}
 \end{flushleft}
 \end{minipage}
 \begin{minipage}{0.333\textwidth}
 \begin{flushleft}
 \includegraphics[width=0.92\textwidth]{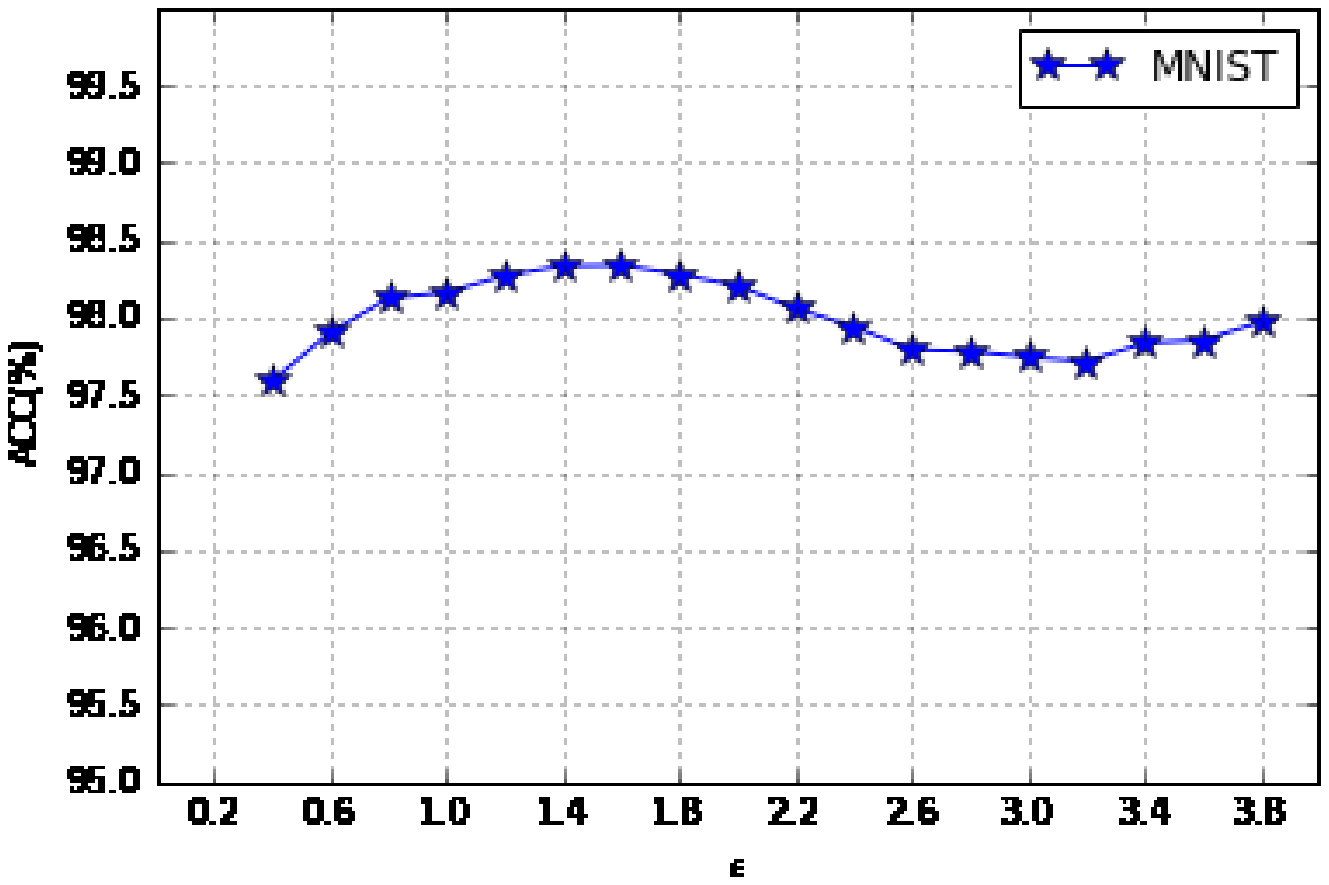}
 \caption{\footnotesize Impact of parameter $\epsilon$.} 
 \label{fig:epsilon}
 \end{flushleft}
 \end{minipage}
\begin{minipage}{0.334\textwidth}
 \begin{flushleft}
 \includegraphics[width=1\textwidth]{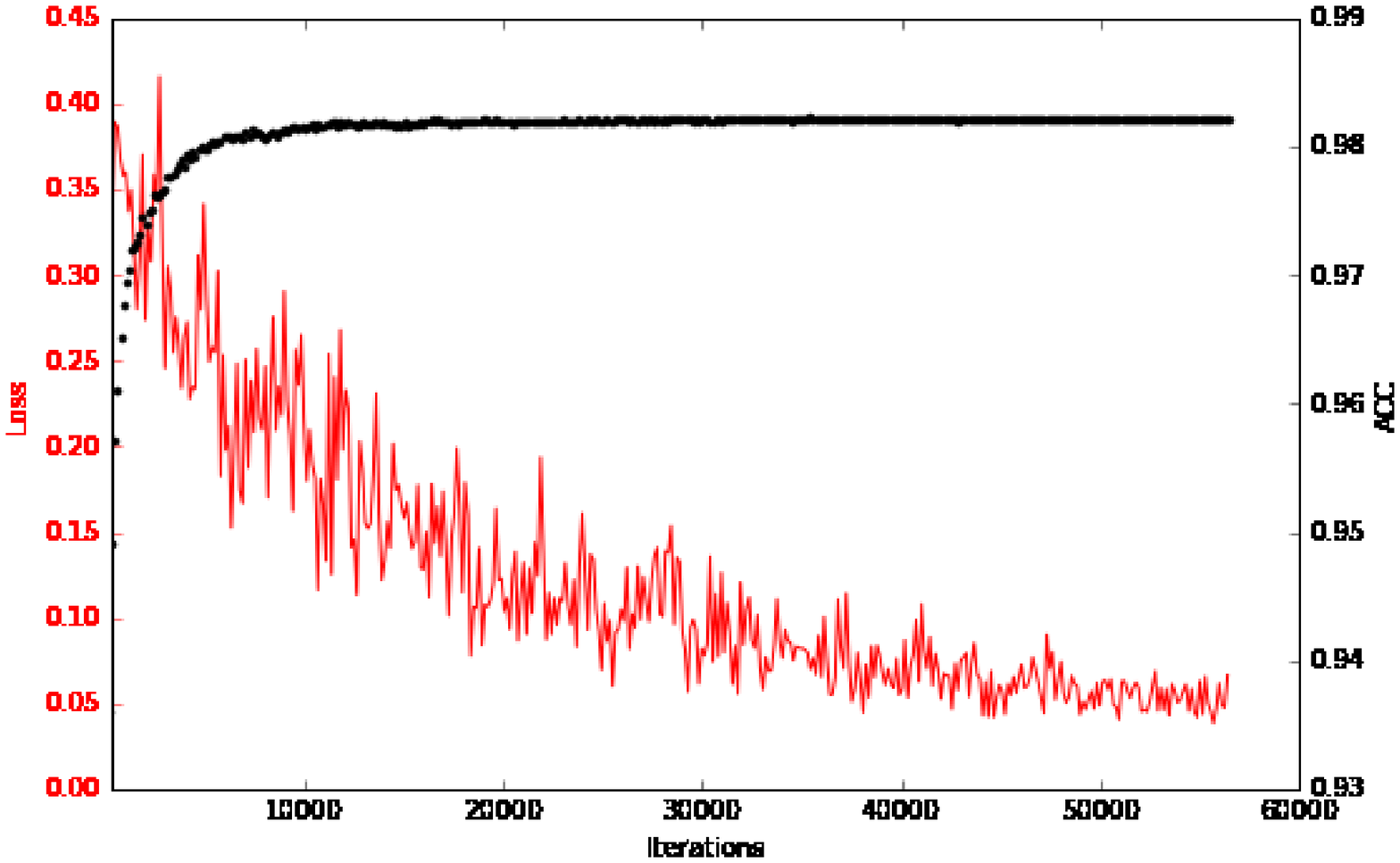}
 \caption{\footnotesize Convergence of RDEC.} 
 \label{fig:convergence}
 \end{flushleft}
 \end{minipage}
 \end{tabular}
 \end{figure}

Parameter $\gamma$ has an adjusting leverage function in balancing two terms of the objective function Eq.~(\ref{eq:loss}). When $\gamma$ is small (resp., large),  the effect of $L_V$ is small (resp., large) and the effect of  $L_D$ is large (resp., small). This effect was investigated by executing RDEC on MNIST and imbalanced MNIST with $r_{min}=0.1$ under different settings. All experiments were executed five times, the mean and standard deviation of accuracies were shown in Figure \ref{fig:gamma}. Overall, RDEC yields stable performance with high accuracies and low deviations on MNIST when $2\leq\gamma\leq14$. 
Accuracies on imbalanced MNIST, by contrast, increase with $\gamma$ when  $2\leq\gamma\leq16$, though the performance becomes unstable when $\gamma\geq7$.
Performance on both datasets suffered obvious degradations at extreme $\gamma$ settings, showing that both the $L_D$ and $L_V$ terms are important for RDEC. 
We recommend setting $\gamma$ between $2$ and $6$.
 
.
\subsection{Impact of perturbation size $\epsilon$}
\label{impactEpsilon}
 Perturbation size $\epsilon$ specifies the range of nearby points requiring consideration during learning. It plays an important role and has been investigated in [\cite{VAT}]. For simplicity, we fixed $\epsilon=1$ based on our experiments on MNIST (Figure \ref{fig:epsilon}). In other specific cases, alternatives such as relative distance from each data to its $i$th-nearest neighbor could be worthy to be considered [\cite{IMSAT}]. 

\subsection{Convergence of RDEC}
According to the objective function in Eq.~(\ref{eq:loss}), RDEC will converge when both its terms are minimized. 
To intuitively demonstrate this convergence, we examined loss and ACC values during the RDEC clustering processes. Figure \ref{fig:convergence} illustrates an example MNIST experiment. In this experiment, $\sigma$ was set to as low as $0.0001$, and as a result clustering stopped after about $56,000$ iterations. From this figure we can see a gradual decline in loss throughout the learning process and high, steady performance in ACC after a sharp increase over the first $10,000$ iterations. The loss decline indicates the effectiveness of our optimizing process, and the steady performance of ACC illustrated the reasonability of that clustering procedure stops when the rate of changes in assignments becomes sufficiently small.  

%###############################################----wafer data---- ################################################
\section{Application to defective wafer map detection }
\label{waferExperiments}
We adopt RDEC to the detection of defect patterns in wafer maps produced in semiconductor manufacturing. Wafer defect maps show spatial patterns of defective chips on wafers. Characteristic patterns in defect maps found in manufacturing test results suggest crucial trouble occurring somewhere in the fabrication processes. 
Early detection of such patterns is thus essential for yield improvement.% in semiconductor manufacturing. 

To reduce the time and labor costs of visually checking all wafer maps, previous works [\cite{liao2014similarity}][\cite{nakata2017comprehensive}] have investigated automated classification with clustering. However, these works do not address the imbalanced-data problem, which is inevitable because there are usually far fewer maps showing defect patterns than not.
In this work, real-world data collected in semiconductor fabrication are used to evaluate the performance of RDEC.

\subsection{Dataset and experimental settings}
\begin{figure}[htp]
\begin{center} 
\includegraphics[width=0.9\textwidth]{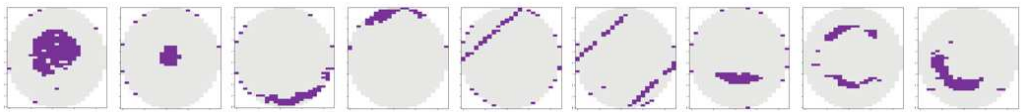}
\caption{Examples of wafer defect maps.}\label{fig:waferMaps}
\end{center}
\end{figure}

A wafer defect map is represented as a two-dimensional binarized image. Figure \ref{fig:waferMaps} shows some example defect wafer maps. Each circle corresponds to one wafer, and colors indicate whether the manufactured chip is defective (purple) or non-defective (gray). As the figure shows, defective chips can appear in characteristic patterns, such as large central circles and linear scratches. In this work, each wafer defect map is converted into a binary vector with dimension equal to the number of chips manufactured on the wafer.

In this experiment, defective maps are selected from 11 classes. For evaluation, we visually checked maps and prepared 11 labels based on their defect patterns, such as CENTER\_DOT, TOP\_EDGE, BOTTOM\_EDGE, UPPER\_LEFT\_SCRATCH, or BOTTOM\_MIDDLE. We labeled about $13,000$ wafers sampled from the results of quality testing in semiconductor fabrication. We evaluated the performance of clustering methods with ACC as in Section \ref{Experiments}. 
Although the number of class is not known in advance in reality, we set the number of clustering to the optimal value of 11 in this experiment.

As the number of similar maps varies depending on the magnitude of the trouble, label distributions in the wafer dataset are highly imbalanced. Table \ref{tab:Ratio} shows the ratio of each label in the raw wafer dataset. The majority class covers $68.57\%$ of whole dataset, while five minor classes cover less than 1.0\% of wafers. The $r_{min}$ of $0.0006$ is extremely low, but such values are not unusual because failures are generally rare in real-world manufacturing.

\begin{table}[htb]
\begin{center}
\caption{Ratio of each class from the whole wafer dataset (\%).}
  \begin{tabular}{c|c|c|c|c|c|c|c|c|c|c|c}\hline
  	  Class&1&2&3&4&5&6&7&8&9&10&11\\\hline\hline
  	  Ratio(\%)&68.57&8.71&7.54&6.02&4.77&2.00&0.99&0.74&0.46&0.12&0.06\\\hline
 \end{tabular}
 \label{tab:Ratio}
 \end{center} 
\end{table}

The network and parameter settings are the same as those for MNIST, except that the update interval $\tau$ was set to $20$, 
the maximum number of iterations $Itr_{max}$ was set to $500$, 
and a convergence constraint was added so that learning stops when the lowest loss of $L_D$ has not been updated in $50$ consecutive iterations. 
This experiment compared RDEC with three methods, K-means, AE+K-means, and DEC.
For pretraining, we utilized a hand-selected dataset consisting of wafer defect maps found by on-site engineers during fabrication. 
In the STL experiment in Section \ref{Experiments}, feature extraction with a network developed by VGG-16 largely improves the performance of clustering, and this result suggests the importance of correct feature extraction. We expect that pretraining with the hand-selected dataset also allows networks to learn appropriate features for wafer dataset to be clustered.

\subsection{Results}
Table \ref{tab:waferAcc} shows clustering results for the wafer dataset. RDEC significantly outperformed the other methods; compared to DEC, RDEC improves accuracy by $37.76\%$.
While DEC, K-means, and AE+K-means divide wafers from the largest class 1 into multiple clusters, RDEC correctly gather most of these into a single cluster, resulting in a high accuracy. 
It is also remarkable that RDEC collects maps from a small class, such as classes 10 and 11, into one cluster. Since finding small but distinctive clusters is essential to early detection of patterns in wafer defect maps, RDEC have a competitive advantage in this application. 

\begin{table}[htb]
\begin{center}
\caption{Clustering accuracy on wafer dataset (\%).}
  \begin{tabular}{c|c|c|c|c}\hline
  Method	&K-means&AE+K-means&DEC&RDEC\\\hline\hline
  ACC&38.00&41.95&42.10&79.86\\\hline
 \end{tabular}
 \label{tab:waferAcc}
 \end{center} 
\end{table}

\section{Conclusions}
\label{Summary}
This paper proposed RDEC as a method for jointly performing deep clustering and network regularization. 
The data argumentation technique VAT was combined with the DEC in RDEC clustering method.
The effect of this combination was evaluated through analyses and experiments.
By combining VAT with DEC, RDEC alleviated the initial centroid and marginal data problems, and yielded higher performance than current methods on most of datasets. For example, the RDEC accuracy on MNIST was $98.41\%$, which is comparable to results of supervised and semi-supervised learning. 
Particularly, RDEC significantly outperformed other methods on imbalanced datasets. For example, it yielded a high accuracy of $85.45\%$ on a highly imbalanced dataset sampled from MNIST, which is nearly $40\%$ and $8\%$ higher than the accuracy of K-means and DEC, respectively.

\bibliographystyle{plain}
\bibliography{refe}
\end{document}